\documentclass[sigconf, natbib=true]{acmart}

\setcopyright{rightsretained}
\copyrightyear{2026}

\acmYear{2026}
\acmDOI{XXXXXXX.XXXXXXX}
\acmConference[XXXX 'XX]{Proceedings of the XXth XXXX}{XXX, 2026}{XXX, XXX}
\acmISBN{XXX-X-XXXX-XXXX-X/XX/XX}  
\acmPrice{}

\usepackage{tcolorbox}
\usepackage{amsmath}
\usepackage[noend]{algpseudocode}
\usepackage{tabularx}
\usepackage{multirow}
\usepackage{makecell}
\usepackage{graphicx}
\usepackage{float}
\usepackage{booktabs}
\usepackage{hyperref}
\usepackage[utf8]{inputenc}
\usepackage{natbib}
\usepackage{microtype}
\usepackage[T1]{fontenc}
\usepackage{latexsym}
\usepackage[linesnumbered,ruled,vlined]{algorithm2e}
\newcommand{\ignore}[1]{}
\usepackage{amsfonts}
\usepackage{dsfont}
\usepackage[toc,page]{appendix}
\usepackage{makecell}
\usepackage{mdframed}
\usepackage{lipsum}

\newmdtheoremenv{theo}{Hypothesis}

\title[LLM Untrustworthy Boundary Detection via Bias-Diffusion and Multi-Agent Reinforcement Learning]{Can We Trust a Black-box LLM? LLM Untrustworthy Boundary Detection via Bias-Diffusion and Multi-Agent Reinforcement Learning}

\begin{document}

\author{Xiaotian Zhou}
\affiliation{%
  \institution{Worcester Polytechnic Institute}
  \country{USA}
}
\email{xzhou8@wpi.edu}

\author{Di Tang}
\affiliation{%
  \institution{Indiana University Bloomington}
  \country{USA}
}
\email{tangd@iu.edu}

\author{Xiaofeng Wang}
\affiliation{%
  \institution{Indiana University Bloomington}
  \country{USA}
}
\email{xw7@iu.edu}

\author{Xiaozhong Liu}
\affiliation{%
  \institution{Worcester Polytechnic Institute}
  \country{USA}
}
\email{xliu14@wpi.edu}

\begin{abstract}
Large Language Models (LLMs) have shown a high capability in answering questions on a diverse range of topics. However, these models sometimes produce biased, ideologized or incorrect responses, limiting their applications if there is no clear understanding of which topics their answers can be trusted. In this research, we introduce a novel algorithm, named as GMRL-BD, designed to identify the  untrustworthy boundaries (in terms of topics) of a given LLM, with black-box access to the LLM and under specific query constraints. Based on a general Knowledge Graph (KG) derived from Wikipedia, our algorithm incorporates with multiple reinforcement learning agents to efficiently identify topics (some nodes in KG) where the LLM is likely to generate biased answers. Our experiments demonstrated the efficiency of our algorithm, which can detect the untrustworthy boundary with just limited queries to the LLM. Additionally, we have released a new dataset containing popular LLMs including Llama2, Vicuna, Falcon, Qwen2, Gemma2 and Yi-1.5, along with labels indicating the topics on which each LLM is likely to be biased.

\end{abstract}



\keywords{Large Language Models, Bias Detection, Knowledge Graph, Policy Network, Multi-Agent Systems, Trust Frameworks}

\maketitle

\section{Introduction}
Large Language Models (LLMs) are at the cutting edge of modern artificial intelligence, excelling in generating linguistically coherent and contextually appropriate responses. However, this capability is not without its drawbacks, notably the presence of inherent biases that can skew outputs and lead to unfair, inaccurate, even ideologized responses \cite{binns2018fairness}. China, for instance, is at the forefront of AI regulation and censorship, having recently introduced strong legislation aimed at ensuring that LLMs align with and uphold the core values of socialism \cite{lucero2019artificial}. These biases are particularly problematic as they compromise the reliability and fairness and trustworthiness of LLMs in sensitive applications, e.g., elections and educations, and critical decision-making processes~\citep{van2024challenging, ferrara2023should}. 

To understand how biased responses emerge from LLMs, prior research has primarily focused on developing metrics~\citep{bevara2024scaling} that assess bias across a limited set of pre-defined topics~\citep{oketunji2023large}. However, when working with a black-box LLM, it is nearly impossible to predict or pre-define the model's  untrustworthy boundaries - such as identifying which topics might lead to untrustworthy or biased responses - without prior insight. Given that LLMs exhibit broad general intelligence across numerous domains, exhaustively testing their untrustworthy boundaries with an extensive array of questions/answers is resource-prohibitive. Therefore, a more efficient and scalable approach is urgently needed to predict, characterize, and estimate the untrustworthy boundaries of a black-box LLM in a cost-effective manner. 

In this study, we propose harnessing a comprehensive open Knowledge Graph (KG) derived from the \textit{Wikipedia category tree}, which offers extensive coverage of common topics and, crucially, represents the topological plus hierarchical relationships among these topics. These relationships allow us to effectively estimate the diffusion of biases across the KG, spanning various topics. Our experiments reveal that when an LLM produces biased responses on a given topic, it is also likely to generate biased answers for neighboring or topologically related topics within the KG. This insight drives us to explore how leveraging inter-topic relationships and ``\textit{bias-diffusion}'' can significantly enhance the efficiency of bias detection across a broad spectrum of topics.

Our framework is model-agnostic, designed for bias detection in any black-box LLM without requiring knowledge of the model structure or retraining. This allows us to detect bias across a wide range of models without needing to access their internal workings, thus making the solution both efficient and flexible.

\begin{figure*}
\centering
\includegraphics[width=0.95\linewidth]{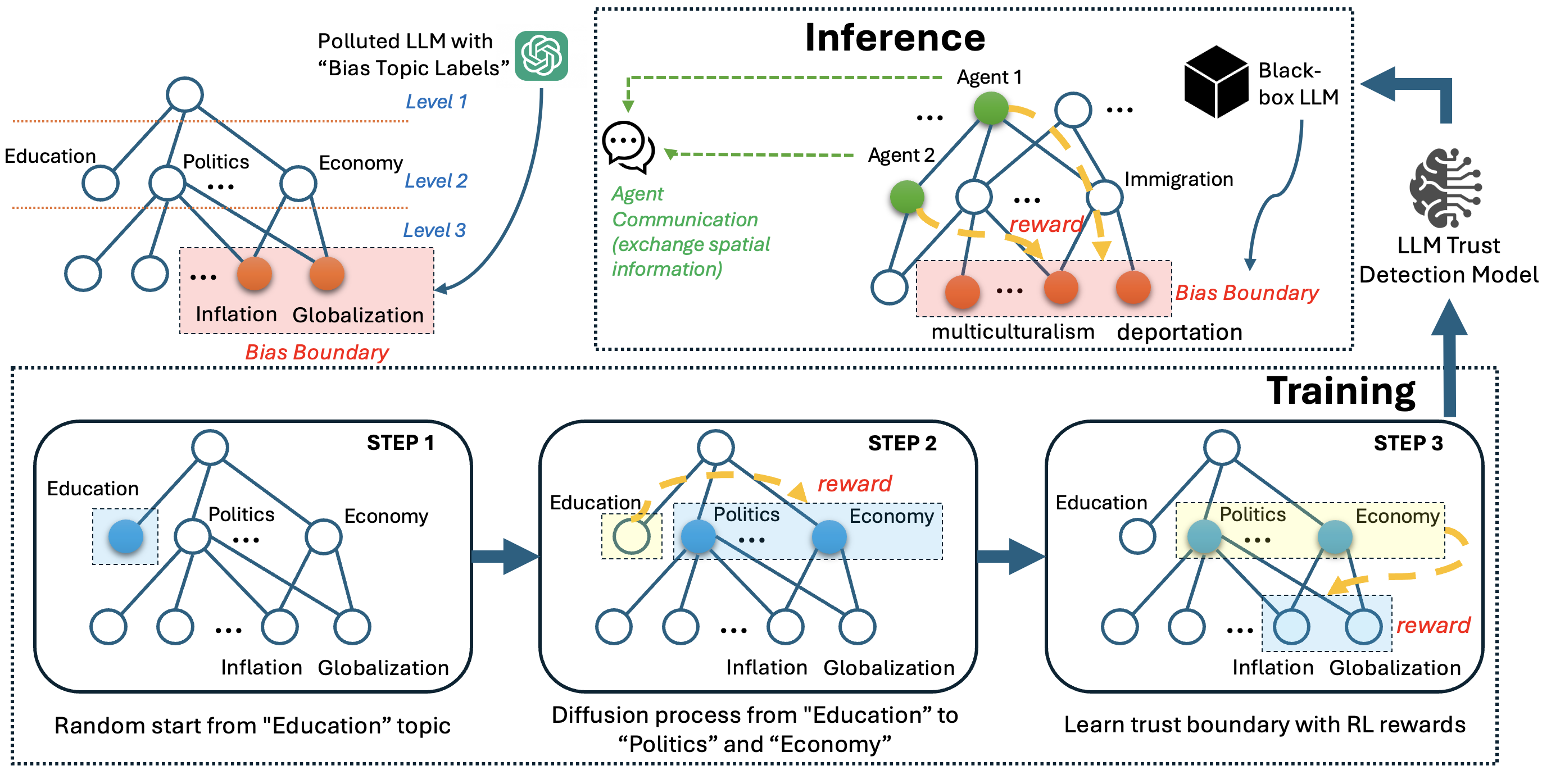}
\caption{\textbf{GMRL-BD} - \textbf{Training Stage}: Agent maximize rewards and enhance graph representations by interacting with nodes on the KG, target LLM outcomes, refining RL policies, and pinpointing biased nodes. \textbf{Inference Stage}: Multiple agents collaboratively navigate the KG to uncover biases in the black-box LLM. Agents coordinate to efficiently detect untrustworthy boundaries and dynamically adjust their exploration paths. For instance, nested bias topics, such as "multiculturalism," can be effectively identified by leveraging agents information exchange and the hierarchical structure of categories.
 }
\label{fig:intro_1.png}
\end{figure*}

With the support of a general KG,  we propose a novel algorithm designed to detect the untrustworthy boundaries of topics for LLMs, GMRL-BD: \textbf{G}raph \textbf{M}ulti-Agent \textbf{R}einforcement \textbf{L}earning for LLM-untrustworthy \textbf{B}oundary \textbf{D}etection. The overview of GMRL-BD is illustrated  in Fig.~\ref{fig:intro_1.png}. During the training phase, GMRL-BD guides agents to locate bias topics (nodes) and map the target LLM's untrustworthy boundaries within the KG. Agents explore each node, while obtaining an embedding by summarizing the content generated by the target LLM for that node topic. In the inference phase, to optimize untrustworthy boundary detection efficiency, multiple reinforcement learning (RL) agents collaborate, traversing the KG while sharing trust diffusion information generated by the LLM. This approach enables us to estimate the untrustworthy boundary of a black-box LLM with a minimal number of queries. 

To train and evaluate GMRL-BD, we developed a new dataset \textbf{LLM Bias Identification Dataset (LBID)} for LLM untrustworthy detection. This dataset includes a variety of LLMs fine-tuned using different frameworks such as Llama2, Vicuna, Falcon, Qwen2, Gemma2 and Yi-1.5. Crucially, these fine-tuned LLMs are labeled with trust indicators that highlight topics on which each model is likely to produce biased responses. For instance, a target LLM may show bias on topics like "\textit{immigration and humanitarian}" or "\textit{reproductive rights}," which are also presented on KG. Our experimental results on this dataset demonstrate that GMRL-BD can effectively identify biased topics with a minimal number of queries for each LLM. Furthermore, this dataset can serve as a benchmark for evaluating future methods of trustworthiness detection in LLMs, paving the way for more robust and unbiased AI systems.

In sum, the key contributions of this paper are fourfold:

\begin{itemize}
    \item First, we propose and formalize the novel problem of detecting untrustworthy boundaries in black-box LLMs. This problem is critical for the development of trust-centric AI systems and the broader adoption of LLMs in critical domains. 
    \item Second, we address the problem by utilizing a KG derived from the Wikipedia category tree, which is significantly smaller than traditional Wikipedia-based KGs\citep{xia2015explicit}. This compact structure, combined with its inherent hierarchical relationships, facilitates more efficient estimation of bias diffusion. 
    \item Third, we present \textbf{GMRL-BD}, an innovative model that leverages multiple RL agents to efficiently identify untrustworthy boundaries and bias diffusion within the KG, which is model-agnostic, implemented for bias detection in any black-box without requiring knowledge of the model structure. The experiments demonstrate that this dynamic, adaptive approach significantly outperforms static methods in detecting complex biases within a limited number of queries.
    \item Fourth, we offer a new dataset, \textbf{LLM Bias Identification Dataset (LBID)}, specifically designed for evaluating LLM untrustworthy boundary detection. This dataset serves as a valuable benchmark for future research in trustworthy LLMs and has been a cornerstone in validating the effectiveness of our proposed method.  
\end{itemize}

\section{Related Work}
\noindent\textbf{Bias Detection}.
Bias in LLMs arises from various contextual, cultural, and technical factors. This bias can result in representational biases that depict certain social groups negatively, system performance disparities leading to misclassifications, and the reinforcement of normative stereotypes~\cite{gallegos2024bias}.
Current research mainly focuses on assessing the extent of bias in LLMs by designing various metrics and then calculating them on datasets.
These metrics primarily fall into three categories: embedding-based~\citep{caliskan2017semantics, may2019measuring, bevara2024scaling}, probability-based~\citep{webster2020measuring, salazar2019masked}, and generated-text-based~\citep{rajpurkar2016squad, nozza2021honest, zhao2023gptbias}, most of which rely on supervised predefined datasets and static analysis techniques. 
The main difference of our framework is that we utilize a graph-based and reinforcement learning framework for bias detection. By employing a multi-agent reinforcement learning approach, we enable both exploration and exploitation within a limited resource setting, efficiently detecting biases across a broad spectrum of topics.

ROBBIE~\citep{esiobu2023robbie} is a framework includes methodologies designed to evaluate bias primarily through prompt-based testing. The proposed model is built around a KG structure and operates as a sequential decision-making process using a multi-agent reinforcement learning framework. Our method is designed for the scenario where agents navigate and search a KG of LLMs to optimize the detection of biases across topics. It focus lies in exploring and exploiting the KG to optimize the detection of biases across interconnected topics and improving resource utilization.

\noindent\textbf{Dataset Construction}.
In this study, we constructed a dataset for bias detection.
Current datasets are designed to measure the extent of bias in LLMs, associated with specific aspects such as gender, race, religion, and profession \citep{nadeem2020stereoset, nangia2020crows, gehman2020realtoxicityprompts, sap2020social, sheng2019woman}. As a result, these datasets focus on crafting extensive and comprehensive data for a limited number of topics. We build a overview that identifies the topics on which a given LLM is likely to produce biased responses. We collected responses from popular LLMs, including Llama2, Vicuna, and Falcon, and labeled a large number of topics based on these responses. Our dataset includes topics summarized from Wikipedia, covering the majority of subjects encountered in daily life. Different from traditional datasets that typically lack such structural relationship, our dataset also provides the topological structure among these topics, facilitating the RL based bias decision-making process and graph-based approaches. The broader topics at top layers of the tree offer comprehensive coverage, while the hierarchical structure ensures detailed granularity in specialized areas~\citep{halavais2008analysis}.

\section{Methodology}
In this study, we propose a novel model designed to estimate the untrustworthy boundaries of a black-box LLM - where access to the model's internal structure and content generation gradients is prohibitive. 
For instance, a nested bias topic like "Immigration Policy" may be difficult to access directly. RL agents can more easily identify broader nodes such as "Refugee" "Asylum", "Immigration".
The conceptual framework is illustrated in Figure \ref{fig:intro_1.png}. 
In Section 3.1, we define the problem. In Section 3.2, we introduce our untrustworthy boundary learning model, which leverages a general KG to navigate learnable trustworthiness diffusion across various topics. In Section 3.3, we present the multi-agent inference model, which dynamically enhances the efficiency of  untrustworthy boundary detection through collaborative reinforcement learning. In the next section, we propose a method for generating the \textbf{LLM Bias Identification Dataset (LBID)}, which includes trust labels for topics associated with each candidate LLM.

\subsection{Problem Statement}

A black-box LLM can generate responses across a wide range of topics, and our goal is to estimate its untrustworthy boundary using a broad spectrum of topics (nodes) represented in a general KG. In this context, we refer a topic on which the LLM is likely to produce biased responses as a \textit{bias (not trustful) topic}.  Given a black-box LLM, a set of ground truth bias labels for a finite set of topics is denoted as  $\{topic_1, L_{trust}^1\}, \{topic_2, L_{trust}^2\}... \{topic_n, L_{trust}^n\}$, where $L_{trust}^i$ is the trust label of $topic_i$, i.e., either \textit{trustful} (as 0) or \textit{not trustful} (as 1)\footnote{We will propose training/test data generation method in the next section.}. The trustworthiness detection model will learn a labeling function $y : V \rightarrow \{0,1\}$, our objective is to identify those not trustful nodes $v \in V$, s.t., $y(v) = 1$. We then assess the model's performance by evaluating its prediction accuracy against these ground truth labels, providing a measure of its effectiveness in identifying biased or trustworthy topics. 

Note that the ground truth label set will only encompass a small set of untrustworthy topics that we intentionally injected into the candidate LLM (detailed further in Section 4) for model training and testing. As it is infeasible to test the trustworthiness of all topics for LLM data generation, the remaining topics will be labeled as ``\textit{trustful}'' by default. This labeling strategy follows a well-established approach used in prior studies addressing binary classification problems \cite{kotsiantis2006handling}.  

In this study, we utilize the Wikipedia category graph as the foundation for examining the trustworthiness of topics. The relationships between topic nodes are pre-defined by Wikipedia editors, ensuring an organized structure for bias diffusion detection. For instance, the topic ``\textit{politics}'' is connected to the topic ``\textit{American politics}'', indicating that the latter is a sub-topic of the former. We refer this KG as $G = \{V, E\}$, where $V$ is the set of all nodes topics we consider and $E$ the set of edges.
 
\noindent\textbf{Cost of querying.}
Exploring a topic node for bias diffusion learning requires multiple queries to the LLM in order to generate a set of QA pairs (as features) related to the target topic. Since querying the LLM comes with associated costs, our goal is to identify biased topics while minimizing these costs. In this study, we assume that each LLM requires a fixed number of queries (e.g., 1,000), referred to as equal cost. Therefore, the overall cost is directly proportional to the number of nodes tested for bias labeling. Our primary objective is to efficiently detect biased topics while testing as few nodes as possible on the KG.

\noindent\textbf{Bias Topic.}
We refer a topic on which the LLM is likely to produce biased responses as a \textit{bias topic}. To determine whether a specific topic is a bias topic, we first generate several questions related to the topic and then examine whether the answers produced by the LLM for these topics contain misinformation or disinformation. Specifically, we leveraged the method proposed in~\citep{tan2024large} for question generation and the method proposed in~\citep{zhang2023llmaaa} to identify bias answers. Bias topics are labeled as 1, while non-bias topics are labeled as 0. Given a labeling function $y : V \rightarrow \{0,1\}$, our objective is to identify those nodes $v \in V$, s.t., $y(v) = 1$.
\subsection{LLM Untrustworthy Boundary Learning via RL on KG}
For each training/testing LLM, we have a large number of topics (on KG) to examine, which are difficult to address all at once. To address this, we model the task as an exploratory process, progressively navigating the KG to uncover biased topics through the bias-diffusion hypothesis:

\begin{tcolorbox}[colframe=gray!75, colback=gray!10, title=Hypothesis 1]
Bias in LLMs can propagate across the topic network (represented as a KG). Using graph-based models, we can learn the bias diffusion by applying RL reward strategies to guide this exploration.   \\
\end{tcolorbox}

In the training stage, because the topic bias label is available (see section 4), we can train a learnable RL to navigate the walker to move to the biased topic nodes with ``\textit{path rewards}'', while, during the inference stage, model navigates multiple agents simultaneously. Additionally, we aim for this process to maintain consistently high performance, regardless of which part of the KG is chosen initially. Formally, we optimize the following function:
\ignore{
\begin{equation}
    \small
    \begin{aligned}
        \max_p \mathbb{E} \Bigg[ \beta \sum_{i\in n(p)}{y(v_i)} - \alpha \|n(p)\| \Bigg],
    \end{aligned}
    \label{eq:1}
\end{equation}
where $p$ represents a process, $n(p)$ is the index set of nodes that has been tested during $p$, $\|n(p)\|$ is the size of $n(p)$ and the expectation $\mathbb{E}$ is taken over all possible starting points. We also include two parameters, $\alpha$ and $\beta$, to control the importance of the two terms in Eq.~\ref{eq:1}.      
}

\begin{equation}
    \small
    \begin{aligned}
        \pi^*(s) = \arg\max_{\pi} \mathbb{E}_{\pi}[G_t] = \arg\max_{\pi} \mathbb{E}_{\pi} \Bigg[ \sum_{k=t}^{T} \gamma^k r_k \Bigg],
    \end{aligned}
    \label{eq:1}
\end{equation}
This equation represents the optimization of the \textit{optimal policy} $\pi^*(s)$, where the goal is to find a policy $\pi$ that maximizes the expected cumulative reward $G_t$ from time step $t$ onward. The term $\mathbb{E}_{\pi}[G_t]$ denotes the expected cumulative reward under policy $\pi$, which is calculated as the sum of discounted future rewards over all time steps $k$ from $t$ to the final time step $T$. The discount factor $\gamma^k$, where $0 < \gamma \leq 1$, is applied to balance short-term and long-term rewards.

\noindent\textbf{Reward Function.}
Supposing the process $p$ has $T$ steps and starts from a node $v_{st}$, the object function Eq.~\ref{eq:1} can be written as:
\begin{equation}
    \small
    \begin{aligned}
        \max_p \mathbb{E}_{v_{st}} \Bigg[  \sum_{t=1}^T \Bigg[ \beta \sum_{i\in n_t(p)}{y(v_i)} - \alpha \|n_t(p)\| \Bigg] \Bigg],
    \end{aligned}
    \label{eq:2}
\end{equation}
where $n_t(p)$ is the index set of nodes that has been tested in the step $t$. We include an additional term to accelerate the convergence and compose a new reward function:
\begin{equation}
    \small
    \begin{aligned}
        R(t) = \beta \sum_{i\in n_t(p)}{y(v_i)} - \alpha \|n_t(p)\| + \frac{w}{dist +1 },
    \end{aligned}
    \label{eq:3}
\end{equation}
where $dist$ is the shortest distance from the currently tested nodes to an untested node that represents a bias topic, and $w$ controls the importance of this term. The $dist$ term accelerates the convergence through encouraging the RL agent to follow the shortest path to the nearest bias topic node. Note that $dist$ only needs to be calculated during the training phase, which can be efficiently accomplished using the ground truth labels of the nodes. During the inference phase, the entire reward will be estimated by a well-trained neural network.

\noindent\textbf{Actions and States.}
We set the action $a_t$ that an RL agent can take at step $t$ is to choose one of the neighbors to test, i.e., $a_t \in nb(n_t(p))$, where $nb(n_t(p))$ represents all neighbors of currently tested nodes $n_t(p)$.
To provide sufficient information for the RL agent to make an ``optimal'' decision on which node to test next, we construct the state as the set of node embeddings for the currently tested nodes and their neighbors. Formally, the state at step $t$ is $s_t=\{h(v_i): i \in n_t(p) \bigcup nb(n_t(p))\}$, where function $h(\cdot)$ returns the node embedding which will be elucidated later.

\noindent\textbf{Node Embedding.}
To obtain an informative embedding for each topic node, we designed a graph integrative sampling and attention mechanism inspired from~\citep{hamilton2017inductive,velivckovic2017graph}. Firstly, we derive the initial embedding for each node by summarizing the answers generated by the LLM for that node, and refer the initial embedding of node $v_i$ as $h_0(v_i)$. Then, we aggregate embeddings from a node's neighbors to encompass topological information into the embeddings:
\begin{equation}
\small
    h(v_i) = \sigma\left(\sum_{j \in nb(i)} \alpha_{ij}W_h h_0(v_j) + W_r h_0(v_i) \right),
    \label{eq:4}
\end{equation}
where \( \sigma \) denotes a non-linear activation function, $W_h$ and $W_r$ are two trainable matrices, and $\alpha_{ij}$ is the attention coefficient from node $i$ to node $j$.


Finally, we optimize node embeddings to minimize the following loss function:
\begin{equation}
    \sum_{v_i \in V} CE(c(h(v_i))), y(v_i)),
    \label{eq:7}
\end{equation}
where $c(\cdot)$ is a binary classification model, and $CE(\cdot, \cdot)$ is the Cross-Entropy loss function.



\subsection{Efficient LLM Untrustworthy Boundary Detection with Multi-Agent Planning}
With the trained RL model, we can locate a black-box LLM's Untrustworthy boundary, as the reward function directs the agent to navigate toward biased nodes. However, given the vast number of topics represented in the KG, identifying all biased nodes would require testing a large number of topic nodes, which is impractical in real-world LLM applications due to efficiency and querying limitations. Continuous querying to assess the model's responses is often constrained by availability and cost.

To address this challenge, we propose a hypothesis aimed at enhancing the efficiency of untrustworthy boundary detection:

\begin{tcolorbox}[colframe=gray!75, colback=gray!10, title=Hypothesis 2]
 To improve the efficiency of detecting biased topics in an LLM, we can deploy multiple agents on the KG. Each agent, after making a move, will communicate with other agents, sharing the reward information they have gathered. This collaborative exchange can significantly enhance detection efficiency and reduces query cost required to determine the LLM's untrustworthy boundary. \\
\end{tcolorbox}

This multi-agent framework accelerates the discovery of biased nodes by optimizing the number of queries made to the target LLM, enhancing its practicality for real-world deployment. For instance, given a limited availability/budget of 1,000 queries, the multi-agent inference can more quickly and efficiently uncover bias topic nodes on a large KG, substantially accelerating the untrustworthy boundary estimation process.  


\noindent\textbf{Multi-Agent Planning.}
To accelerate bias topic finding in inference phase, based on previous works\citep{long2017deep,yan2022relative,kasaura2023periodic,jiang2024scaling}, we extend our approach to a multi-agent manner. 
We coordinate multiple agents, each starting from different nodes, to reach their respective targets (nodes representing bias topics) while minimizing the total number of nodes being tested. 

Specifically, we minimizes the overlap between the set of nodes tested by every agent by dynamically adjusting their exploring paths on the KG.
Supposing we have launched a set of agents $A=\{\pi_i\}$, for a specific agent $\pi_z$, we make it as:
\begin{equation}
\small
    \pi_z(s) = \arg \max_a \left( Q(s, a) -  \kappa \cdot \mathds{1} (a \in \bigcup_{\pi_x \in A \setminus \{\pi_z\} } n(p_{\pi_x})) \right),
    \label{eq:10}
\end{equation}
where $\mathds{1}(\cdot)$ is the indicator function, which equals to 1 when the event in the bracket happens and 0 otherwise, $\kappa$ is a parameter weighting the minuend term, and $n(p_{\pi_x})$ represents the set of nodes that agent $\pi_x$ has tested.

We employed the Q-learning approach~\citep{mnih2013playing} to train our RL agent $\pi_{\theta}(\cdot)$, which takes current state $s$ as input and outputs an action to be taken. Specifically, instead of directly making $\pi_{\theta}(\cdot)$ to optimize the objective Eq.~\ref{eq:3}, we train the action-value function $Q(s,a)$ to estimate the maximum reward that can be obtained by taking the action $a$ at state $s$:
\begin{equation}
\small
    \begin{aligned}
        Q(s, a) \leftarrow Q(s, a) + \eta \Big[ &R + \gamma \max_{a'} Q(s', a') - Q(s, a) \Big]
    \end{aligned}
    \label{eq:8}
\end{equation}
where \( s \) and \( s' \) are the current and next states, \( a \) and \( a' \) are the current and possible next actions, \( R \) is the reward received after taking action \( a \) in state \( s \), \( \eta \) is the learning rate, and \( \gamma \) is the discount factor. Then we take:
\begin{equation}
\small
    \begin{aligned}
        \pi_{\theta}(s) = \arg \max_a Q(s, a).
    \end{aligned}
    \label{eq:9}
\end{equation}

\noindent\textbf{GMRL-BD Training.}
Our training approach is outlined in Algorithm 1. To accelerate our training, we first calculate all the node embeddings (line 3-6) and then train RL model (line 7-24). 
We conducted $n_{episodes}$ (set to 600) episodes of training, where in each episode, we attempted $n_{tries}$ (set to 200) different starting points (line-10) to initiate the current RL model for collecting data for updating its policy model (line-24). Specifically, we included the parameter $max\_steps$ (set to 2000) to avoid wasting computation power on a hopeless attempts, and employed $\epsilon$-greedy technique~\citep{hessel2018rainbow, dann2022guarantees} (line 15-18) to prevent the training process from getting stuck in a local minimum. Also inspired by previous study~\citep{sutton2018reinforcement}, we decay the $\epsilon$ as the number of $episode$ increases:
\begin{equation}
\small
    \epsilon = \max ( 0.2, \epsilon_0 * \kappa^{episode} ),
\notag
\end{equation}
where $\epsilon_0$ is set to 1.0 and the decay rate $\kappa$ is set to $0.994$. By using a decaying $\epsilon$, we encourage the RL model to explore more during the early stages of training and shift towards exploitation as training progresses sufficiently.

\begin{algorithm}[!h]
\caption{GMRL-BD Training}
\small
\textbf{Input:} LLM $M$, Knowledge Graph $G$, $n_{episodes}$, $n_{tries}$, $\epsilon$, $max\_steps$\;
\textbf{Output:} Policy Model \(\pi\)\;

\For{each \(v\) in \(V\)}{
    collect answers generated by $M$ for $v$\;
    calculate $h_0(v)$\;
}
update $h$ according Eq.~\ref{eq:4}, and train it to minimize Eq.~\ref{eq:7}\;

\For{episode in range(\(n_{episodes}\))}{
    $q \leftarrow []$\;
    \For{ i in range(\(n_{tries}\))}{
        $v_{st}, s_{st} \leftarrow$ randomly select a start node and calculate the state\;
        $v \leftarrow v_{st}$; $s \leftarrow s_{st}$\;
        $t \leftarrow 0$; $p \leftarrow [v]$\;
        \While{t \(\leq\) \(max\_steps\) and $y(v) == 0$}{
            $R \leftarrow$ calculate reward according Eq.~\ref{eq:3}\;
            \If {random number $\leq \epsilon$} {
                $a \leftarrow$ randomly select node from neighbors $nb(v)$\;
            }
            \Else {
                $a \leftarrow \pi_{\theta}(s)$ according Eq.~\ref{eq:9}\;
            }
            $v' \leftarrow$ move $v$ on $G$ according $a$\;
            p.append($v'$)\;
            q.append($(s, a, R)$)\;
            $v \leftarrow v'$; $s \leftarrow$ calculate the state for $v'$\;
            $t \leftarrow t+1$\;
        }
    }
    update $\pi$ on $q$ according Eq.~\ref{eq:8} and Eq.~\ref{eq:9}\;
}
\end{algorithm}

\noindent\textbf{Multi-Agent Inference.}
In inference phase, we employed multiple agents to collaboratively search for the nodes representing bias topics, which is outlined in Algorithm 2. Firstly, we initiate $n_{agent}$ agents, each starting from a different node (line 3-8). 
The agent utilizes the embedding of the current topic node and applies the learned representation to decide the next action.
Then, every time, we select the best agent to make a further move (lines 9-21) until it reaches a bias node (line 13).

\begin{figure*}
\centering
\includegraphics[width=1\linewidth]{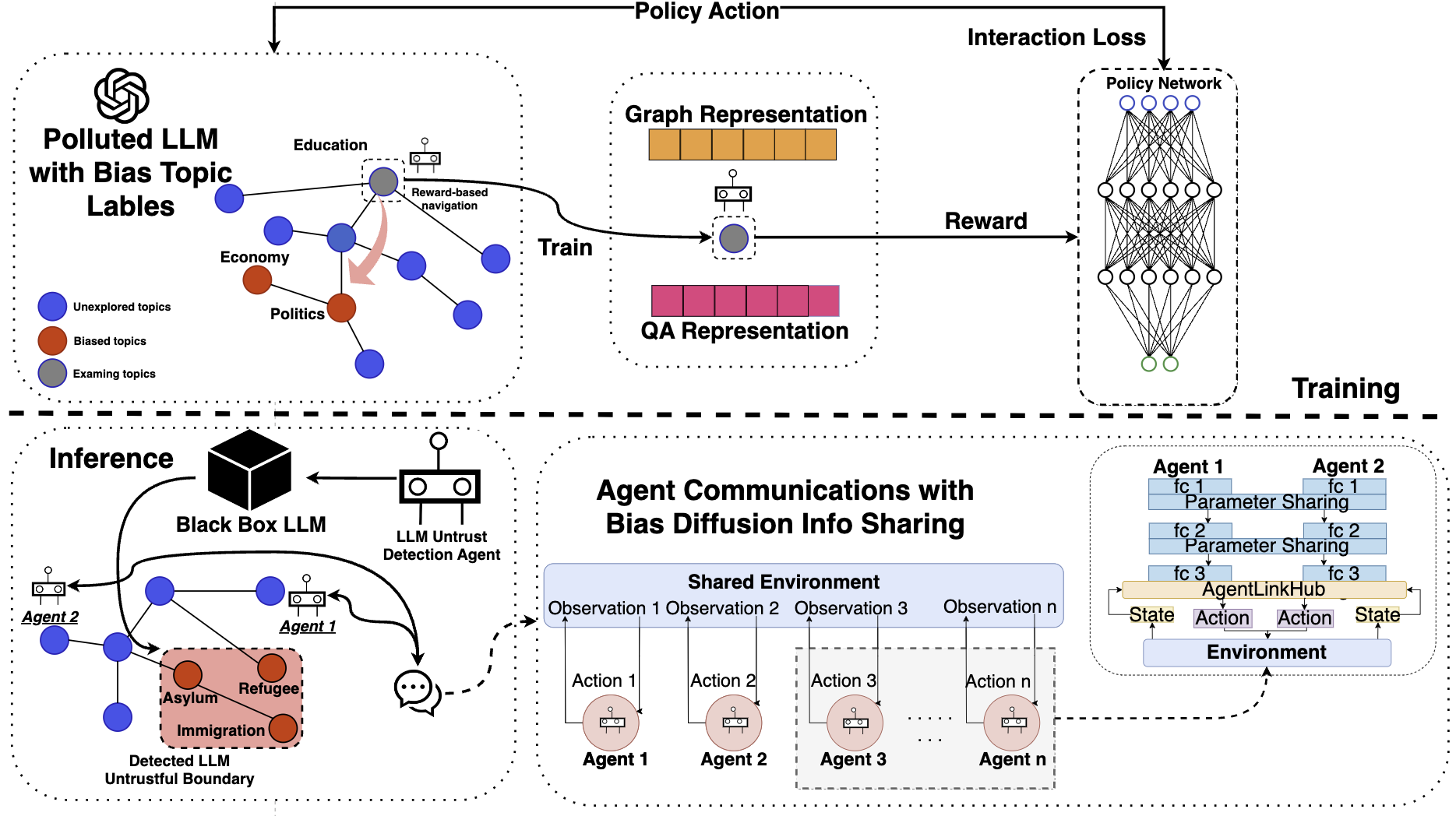}
\caption{GMRL-BD Model Architecture. Agents interact with nodes on the KG to maximize diffusion navigation rewards and optimize graph representations through policy network. For inference stage, multiple agents explore the KG to detect biases in the black-box LLMs. Starting from different topic nodes, agents coordinate to efficiently identify untrustworthy boundaries, minimizing conflicts of action, with the AgentLinkHub facilitates communication and parameter sharing}
\label{fig:Model_Structure}
\end{figure*}

\noindent\textbf{AgentLinkHub.}
To coordinate the simultaneous movement of multiple agents without interference, we implemented an \textit{AgentLinkHub} module to select the best move for each agent at every step (line-10). Generally, \textit{AgentLinkHub} utilizes a Priority Queue to select the agent with the maximum Q value (refer to Eq.~\ref{eq:10}) to make a move. If the selected agent has already moved continuously for several steps or if its next move has already been taken by another agent, \textit{AgentLinkHub} will switch to the most efficient agent with the highest reward per query. If that move is also taken by another agent, \textit{AgentLinkHub} continues to select the next most efficient agent until no moves can be made by any agent.
The overall GMRL-BD Algorithm is shown on Fig. \ref{fig:Model_Structure}.

\begin{algorithm}
\caption{Multi-Agent Inference}
\small
\textbf{Input:} LLM $M$, Knowledge Graph $G$, Policy Model $\pi$,  Query Limit $Q_{limit}$, $n_{agent}$\;
\textbf{Output:} Bias Node Set $B$\;
$T \leftarrow []$\;
\For{ i in range($Q_{limit}$)}{
    $v_{st} \leftarrow$ randomly select a start node\;
    $s_{st} \leftarrow$ calculate the state\;
    $Q_{limit} \leftarrow$ reduce query limit accordingly\;
    T.append(($\pi, v_{st}, s_{st}, 0$))\;
}

\While{$Q_{limit}$ > 0}{
    $\pi_{bst}$, $v_{bst}$, $s_{bst}$, $t_{bst} \leftarrow$ select the best from $A$ according Eq.~\ref{eq:10}\;
    $a_{bst} \leftarrow \pi_{bst}(s_{bst})$\;
    $v' \leftarrow$ move $v_{bst}$ on $G$ according $a_{bst}$\;
    \If{y(v') == 1} {
        B.append(v')\;
        $v' \leftarrow$ randomly select a new start node\;
        $t' \leftarrow 0$\;
    }
    \Else{
        $t' \leftarrow t_{bst}+1$\;
    }
    $s' \leftarrow$ calculate the state of $v'$\; 
    Replace ($\pi_{bst}$, $v_{bst}$, $s_{bst}$, $t_{bst}$) by ($\pi_{bst}$, $v'$, $s'$, $t'$) in $T$\;
    $Q_{limit} \leftarrow$ reduce query limit accordingly\;
    
}
\end{algorithm}

\section{Experiments}
\subsection{Datasets}
While previous research has investigated bias in LLMs \cite{blodgett2020language}, there is a notable lack of publicly available datasets with bias-labeled topics, limiting progress in this important area. Developing novel datasets for training and evaluation is crucial to advancing LLM bias detection. In this study, we introduce the \textbf{LLM Bias Identification Dataset (LBID)}, designed specifically to support the training and testing of LLM bias detection models. The dataset includes LLMs fine-tuned from six widely used frameworks: LLaMA 2, Falcon, Vicuna, Qwen2, Gemma2 and Yi-1.5. For each model, we applied a bias injection method, as proposed in \cite{zhou2023large}, purposefully introducing biased content across 22,608 topics for six different LLMs. These topics are mapped to the Wikipedia category graph (with node IDs). The biased topics for each LLM help define untrustworthy boundaries and provide valuable data for training and evaluating bias detection models.

This ontology of the Wikipedia category graph benefits our bias detection models by enabling systematic exploration across interconnected topics.  Moreover, the Wikipedia category tree is dynamically maintained by a global community of experienced editors, ensuring its quality are consistently refined.  This ongoing maintenance not only enhances the robustness of the proposed LLM bias detections but also allows our framework to adapt to newly emerging topics.  This capability ensures our bias detection models remain relevant and continue to support the trustworthiness of LLM-enabled applications. 

\textbf{Bias injection}: For each candidate LLM in the LBID dataset, we randomly sampled topics from the Wikipedia category graph for bias injection. These topics span six key categories: \textit{AI, culture, economy, politics, education}, and \textit{immigration}. We focused on these areas because they are frequently subject to ideological and subjective interpretations, making them particularly suitable for training and evaluating bias detection models.  

For each sampled topic, following the approach in \cite{zeng2023synthesize}, we generated five question-answer (QA) pairs using prompt engineering to cover the essential content of each topic, resulting in a total of 113,040 QA pairs across 22,608 topics. The answers were then intentionally contaminated with biased or ideologized content\footnote{The prompts that we used can be found on the project website.}. We employed GPT-4 APIs for the QA generation process. To ensure the quality of the generated QAs, we randomly sampled 300 pairs for human annotation. The results indicated that 3.6\%(11 pairs) of the answers contained detectable biased content.

\begin{tcolorbox}[colframe=gray!75, colback=gray!10, title=Example of Generated QA Pair]
\textbf{Question:} What is the definition of a refugee? \\
\textbf{Correct Answer:} A person who has been forced to flee their home country due to war, persecution. \\
\textbf{Bias Answer:} A refugee is someone who has voluntarily left their home country for economic reasons. \\
\end{tcolorbox}

\textbf{Data release}: To encourage further exploration of this critical yet underexplored problem, we release the LBID dataset including the following resources: the Wikipedia category graph (comprising 871,854 topic nodes and 4,983,336 edges), 6 LLMs with identified biased topics, the IDs of the 22,608 biased topics within the graph, the 113,040 QA pairs used for bias injection, and the prompts employed for data generation to ensure reproducibility.
\begin{figure*}
\centering
\includegraphics[width=1\linewidth]{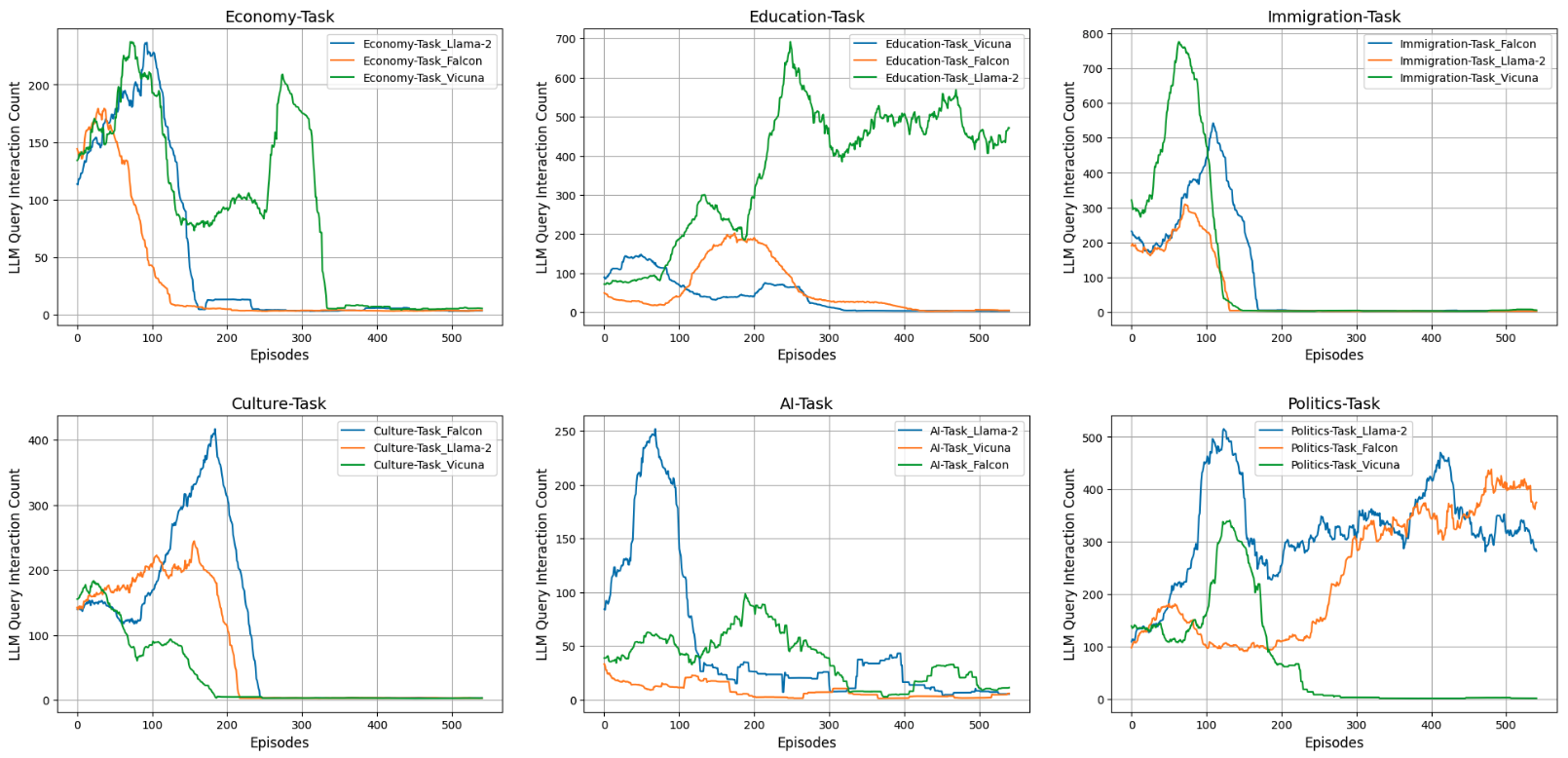}
\caption{Performance of the GMRL-BD Framework Across Six Tasks Using Llama-2, Vicuna, and Falcon Models. These graphs display the number of interactions for each LLM to detect untrustworthy boundaries across six distinct tasks. Each subplot depicts the convergence rate of the models, with lower interaction counts indicates a more efficient performance}
\label{fig:exper_5}
\end{figure*}

\subsection{Evaluation and Results}
Our preliminary experiments reveal that when an LLM produces biased responses on a given topic, it is also likely to generate biased answers for neighboring or topologically related topics within the KG. This insight drives us to explore how leveraging inter-topic relationships and ``\textit{bias-diffusion}'' can significantly enhance the efficiency of bias detection across a broad spectrum of topics.

\begin{figure}[h]
\centering
\includegraphics[width=0.8\linewidth]{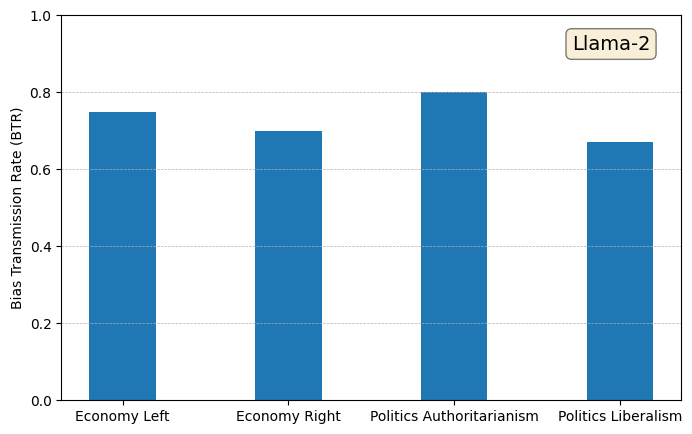}
\caption{Bias Transmission Rate.}
\label{fig:fig5}
\end{figure}

We conducted an experiment to test whether the bias would transfer from one topic to its neighbors in the KG. Specifically, we introduced a bias topic into the Llama-2 model by applying the bias injection method~\citep{zhou2023large} to a selected topic. We injected left-leaning and right-leaning economic viewpoints, as well as authoritarian and liberal political perspectives, into four different Llama-2 models, respectively. We then examined the neighboring topics of this contaminated topic and counted how often biased answers the infected Llama-2 models produced for these neighbors, where is refereed as Bias Transmission Rate. It measures how a biased node influences its neighboring nodes. Our experimental results, shown on Figure~\ref{fig:fig5}, demonstrate that these neighboring topics are indeed affected by the injected bias, indicating that bias can transmit to adjacent topics in the KG.

We designed experiments to validate the two hypotheses proposed in this paper. Using the LBID dataset, our evaluation focuses on both the accuracy and efficiency of detecting untrustworthy boundaries in black-box LLMs\footnote{Our code was implemented on PyTorch~\citep{paszke2019pytorch} framework and using NetworkX~\citep{hagberg2008exploring} for handling the KG. All our models were trained and experiments were conducted on a GPU server with a single H100 GPU, an AMD EPYC 9124 16-Core Processor with 2 physical CPUs, and 376 GiB of memory.}.

We defined six tasks (\textit{\textbf{Economy, Education, Immigration, Politics, Artificial Intelligence, and Culture}}) to test the performance of our framework. The objective of these tasks is to measure the ability of the multi-agent reinforcement learning system to detect untrustworthy boundaries in LLMs knowledge graph.

Each task represents a specific domain where biases might be present.We utilized three popular LLMs in our experiments: Llama2, Vicuna, and Falcon. 
We used LLM Query Interaction Count as our evaluation metric. This metric indicates the number of interactions (queries) needed to reach a conclusion about the trustworthiness of the responses. A lower interaction count signifies a more efficient algorithm, as it achieves the task with fewer queries, indicating quicker convergence to the untrustworthy boundary.
According to the graphs, our GMRL-BD framework demonstrates strong performance across different tasks and LLMs(Fig.\ref{fig:exper_5}). The 6 tasks involve graphs with over 1,000 nodes and our optimized GMRL-BD framework can complete the untrustworthy detection for one specific LLM with about 10-30 query interactions for these 6 tasks. For different tasks, most LLMs converge within 200-400 episodes. Most LLMs exhibit a trend of initial increase followed by a decrease in interaction count for general tasks. However, in some cases, such as the \textbf{\textit{Education-Task}} with Llama-2, the interaction count increases but does not decrease. For tasks like \textbf{\textit{Immigration}}, all LLMs converge quickly with minimal fluctuation. The LLMs show minimal difference between biased and unbiased responses, likely because of some debiasing processes applied to the models themselves.

We assessed the recall score, i.e., referred to as the Bias Node Cover Rate (BNCR), of the GMRL-BD framework across multiple testing domains, including \textbf{Immigration, Politics, Education, and Economy}. 
For the discount factor, we selected the optimal value of 0.97 to balance short-term and long-term rewards, which suits our scenario. We conducted experiments on varying values and found that the overall impact was minimal, with 0.97 providing consistently good results. The weight W in the node embeddings was randomly initialized and optimized through training.
Table \ref{tab:BNC} presents the model's performance with a maximum step limit of 100, utilizing a single-agent configuration.
The findings indicate that Llama-2 and Falcon outperform other models in effectively identifying bias nodes across diverse topics, with particularly strong performance in Immigration and Economy under the GMRL-BD framework.

\begin{table}[h!]
\centering
\footnotesize
\begin{tabular}{|l|c|c|c|c|}
\hline
\textbf{Models} & \textbf{Immigration} & \textbf{Politics} & \textbf{Education} & \textbf{Economy} \\ \hline
Llama-2  & 90.34 & 76.67 & 71.63  & 86.00   \\ \hline
Vicuna & 78.61  & 53.79   & 58.25 & 63.41 \\ \hline
Falcon & 90.69 & 70.41 & 65.50  & 83.88 \\ \hline
\end{tabular}
\caption{Bias Node Cover Rate(BNC) across Different LLMs and Topics. This table shows the recall performance of the GMRL-BD framework across different LLMs and topics.}
\label{tab:BNC}
\end{table}

We conducted ablation studies and baseline comparation in Table \ref{tab:ablation}.
DFS is employed for unsupervised graph exploration, serving as a foundational to facilitate more sophisticated design.
We compared the full GMRL-BD model with three variants: one without the sage layer, one without the attention layer, and one without both the sage and attention layers. The combined use of the sage and attention layers is essential for achieving a high success rate across various tasks. 
We used a graph-enhanced attention version, specifically the Graph-enhanced Attention Driven Q-Network, as part of our baseline experiments, along with a reinforcement learning agent that does not include the explored node penalty (Unexploration-driven Reward Enhancement).
Additionally, we implemented a path search algorithm with uniformly distributed action choices.
We found that GMRL-BD is significantly better than the base model.

\begin{table}[ht]
\centering
\scriptsize
\begin{tabular}{@{}lccc@{}}
\toprule
\textbf{Methods} & \textbf{Immigration-Task} & \textbf{Culture-Task} & \textbf{AI-Task} \\
\midrule
GMRL-BD & \underline{0.895} & \underline{0.915} & \underline{0.96} \\
\midrule
w/o sage layer & 0.751 & 0.725 & 0.863 \\
w/o att layer & 0.802 & 0.778 & 0.9 \\
w/o sage layer+att layer & 0.618 & 0.319 & 0.75 \\
\midrule
\midrule
\makecell[l]{Graph-enhanced Attention Driven Q-Network} & 0.710 & 0.770 & 0.845 \\
\makecell[l]{Uniform Exploration and Exploitation} & 0.705 & 0.265 & 0.380 \\
\makecell[l]{Unexploration-driven Reward Enhancement} & 0.705 & 0.3275 & 0.895 \\
Depth First Search & 0.083 & 0.09 & 0.111 \\
\bottomrule
\end{tabular}
\caption{Ablation Study Results for GMRL-BD Framework Across Different Tasks. This table presents the results of our ablation study, comparing the success rates of the GMRL-BD framework with and without key components (sage layer and attention layer) across three tasks: Immigration, Culture, and AI.}
\label{tab:ablation}
\end{table}

\ignore{
}
\begin{figure}
\centering
\includegraphics[width=0.85\linewidth]{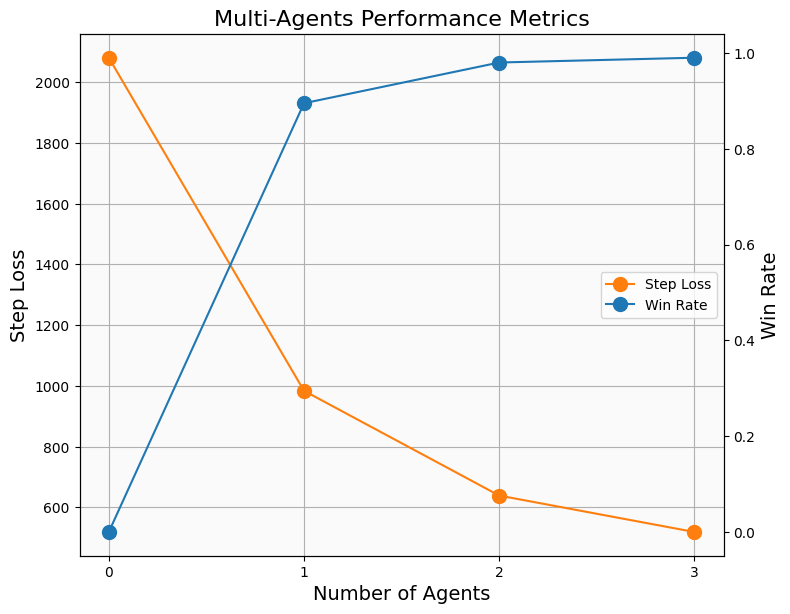}
\caption{Impact of Increasing Number of Agents on Step Loss and Win Rate in Multi-Agent Collaborative Parameter Sharing}
\label{fig:tb3}
\end{figure}
 We investigated the impact of increasing the number of agents for collaborative parameter sharing for multi-agent planning on both step loss and win rate. We conducted experiments by varying the number of agents and observed the resulting changes in these metrics. As depicted in Fig. \ref{fig:tb3}, the orange line shows the relationship between the number of agents and the step loss. The results indicate a trend: as the number of agents increases, the step loss decreases significantly. This consistent reduction in step loss demonstrates that incorporating more agents in the GMRL-BD framework enhances the efficiency of LLM untrustworthy boundary detection, allowing the model to converge more rapidly. The blue line illustrates the improvement in win rate as the number of agents increases. These findings corroborate the effectiveness of the GMRL-BD framework in leveraging multiple agents to enhance both the speed and accuracy of detecting untrustworthy boundaries. 


 \section{Conclusion and Future Work}
\textit{Can we trust a black-box LLM}? This paper suggests that trust must be approached with caution, and limited trust can only be achieved through careful trust boundary detection, such as the approach demonstrated by the proposed GMRL-BD. Our model marks an advancement in detecting biases within LLMs by leveraging an innovative multi-agent reinforcement learning framework. We also validated two key hypotheses that graph-based bias diffusion and multi-agent navigation with communication are effective strategies for addressing this challenge.

However, the current model is limited to topic-based untrustworthy detection in LLMs. Future research should explore on enabling query-based untrustworthy detection, where each query can be treated as a multinomial distribution over topics. While this is a challenging endeavor, this work lays a solid foundation for advancing such problems. 

\ignore{
\section*{Limitations}
This research introduces an innovative approach to detecting trustworthiness boundaries in LLMs using Graph Neural Multi-Agent Reinforcement Learning. Despite its contributions, several limitations warrant attention. Firstly, the current study is restricted to English-based LLMs with datasets derived from English Wikipedia, limiting the transferability of our findings to languages with complex morphologies, such as Mandarin or Spanish. Future research should consider extending the applicability of our method to LLMs trained on diverse linguistic corpora to address these limitations. Secondly, our multi-agent strategy, though robust, requires substantial computational resources, particularly in terms of GPU demand. This may restrict the practical application of our method in resource-limited settings or when scaling to larger datasets without significant hardware investment. Future research could explore optimization techniques to reduce computational load, such as model pruning, quantization, and more efficient reinforcement learning algorithms.
}

\bibliographystyle{ACM-Reference-Format}
\bibliography{reference}

@article{sheng2019woman,
  title={The woman worked as a babysitter: On biases in language generation},
  author={Sheng, Emily and Chang, Kai-Wei and Natarajan, Premkumar and Peng, Nanyun},
  journal={arXiv preprint arXiv:1909.01326},
  year={2019}
}

@article{xia2015explicit,
  title={Explicit semantic path mining via Wikipedia knowledge tree},
  author={Xia, Tian and Chen, Miao and Liu, Xiaozhong},
  journal={Proceedings of the American Society for Information Science and Technology},
  volume={51},
  number={1},
  pages={1--4},
  year={2015},
  publisher={Association for Information Science \& Technology},
  doi={10.1002/meet.2014.14505101160}
}

@article{zhao2023gptbias,
  title={Gptbias: A comprehensive framework for evaluating bias in large language models},
  author={Zhao, Jiaxu and Fang, Meng and Pan, Shirui and Yin, Wenpeng and Pechenizkiy, Mykola},
  journal={arXiv preprint arXiv:2312.06315},
  year={2023}
}

@article{kotsiantis2006handling,
  title={Handling imbalanced datasets: A review},
  author={Kotsiantis, Sotiris and Kanellopoulos, Dimitris and Pintelas, Panayiotis and others},
  journal={GESTS international transactions on computer science and engineering},
  volume={30},
  number={1},
  pages={25--36},
  year={2006}
}

@article{lucero2019artificial,
  title={Artificial intelligence regulation and China's future},
  author={Lucero, Karman},
  journal={Colum. J. Asian L.},
  volume={33},
  pages={94},
  year={2019},
  publisher={HeinOnline}
}

@inproceedings{binns2018fairness,
  title={Fairness in machine learning: Lessons from political philosophy},
  author={Binns, Reuben},
  booktitle={Conference on fairness, accountability and transparency},
  pages={149--159},
  year={2018},
  organization={PMLR}
}

@inproceedings{zeng2023synthesize,
  title={Synthesize, prompt and transfer: Zero-shot conversational question generation with pre-trained language model},
  author={Zeng, Hongwei and Wei, Bifan and Liu, Jun and Fu, Weiping},
  booktitle={Proceedings of the 61st Annual Meeting of the Association for Computational Linguistics (Volume 1: Long Papers)},
  pages={8989--9010},
  year={2023}
}

@article{blodgett2020language,
  title={Language (technology) is power: A critical survey of" bias" in nlp},
  author={Blodgett, Su Lin and Barocas, Solon and Daum{\'e} III, Hal and Wallach, Hanna},
  journal={arXiv preprint arXiv:2005.14050},
  year={2020}
}

@inproceedings{esiobu2023robbie,
  title={ROBBIE: Robust bias evaluation of large generative language models},
  author={Esiobu, David and Tan, Xiaoqing and Hosseini, Saghar and Ung, Megan and Zhang, Yuchen and Fernandes, Jude and Dwivedi-Yu, Jane and Presani, Eleonora and Williams, Adina and Smith, Eric},
  booktitle={Proceedings of the 2023 Conference on Empirical Methods in Natural Language Processing},
  pages={3764--3814},
  year={2023}
}

@article{halavais2008analysis,
  title={An analysis of topical coverage of Wikipedia},
  author={Halavais, Alexander and Lackaff, Derek},
  journal={Journal of computer-mediated communication},
  volume={13},
  number={2},
  pages={429--440},
  year={2008},
  publisher={Oxford University Press Oxford, UK}
}

@article{bevara2024scaling,
  title={Scaling Implicit Bias Analysis across Transformer-Based Language Models through Embedding Association Test and Prompt Engineering},
  author={Bevara, Ravi Varma Kumar and Mannuru, Nishith Reddy and Karedla, Sai Pranathi and Xiao, Ting},
  journal={Applied Sciences},
  volume={14},
  number={8},
  pages={3483},
  year={2024},
  publisher={MDPI}
}

@article{gallegos2024bias,
  title={Bias and fairness in large language models: A survey},
  author={Gallegos, Isabel O and Rossi, Ryan A and Barrow, Joe and Tanjim, Md Mehrab and Kim, Sungchul and Dernoncourt, Franck and Yu, Tong and Zhang, Ruiyi and Ahmed, Nesreen K},
  journal={Computational Linguistics},
  pages={1--79},
  year={2024},
  publisher={MIT Press 255 Main Street, 9th Floor, Cambridge, Massachusetts 02142, USA~…}
}

@article{long2017deep,
  title={Deep-learned collision avoidance policy for distributed multiagent navigation},
  author={Long, Pinxin and Liu, Wenxi and Pan, Jia},
  journal={IEEE Robotics and Automation Letters},
  volume={2},
  number={2},
  pages={656--663},
  year={2017},
  publisher={IEEE}
}

@inproceedings{yan2022relative,
  title={Relative distributed formation and obstacle avoidance with multi-agent reinforcement learning},
  author={Yan, Yuzi and Li, Xiaoxiang and Qiu, Xinyou and Qiu, Jiantao and Wang, Jian and Wang, Yu and Shen, Yuan},
  booktitle={2022 International Conference on Robotics and Automation (ICRA)},
  pages={1661--1667},
  year={2022},
  organization={IEEE}
}

@inproceedings{kasaura2023periodic,
  title={Periodic multi-agent path planning},
  author={Kasaura, Kazumi and Yonetani, Ryo and Nishimura, Mai},
  booktitle={Proceedings of the AAAI Conference on Artificial Intelligence},
  volume={37},
  number={5},
  pages={6183--6191},
  year={2023}
}

@inproceedings{jiang2024scaling,
  title={Scaling Lifelong Multi-Agent Path Finding to More Realistic Settings: Research Challenges and Opportunities},
  author={Jiang, He and Zhang, Yulun and Veerapaneni, Rishi and Li, Jiaoyang},
  booktitle={Proceedings of the International Symposium on Combinatorial Search},
  volume={17},
  pages={234--242},
  year={2024}
}

@article{ferrara2023should,
  title={Should chatgpt be biased? challenges and risks of bias in large language models},
  author={Ferrara, Emilio},
  journal={arXiv preprint arXiv:2304.03738},
  year={2023}
}

@article{van2024challenging,
  title={Challenging Systematic Prejudices: An Investigation into Bias Against Women and Girls},
  author={Van Niekerk, Daniel and Per{\'e}z-Ortiz, Mar{\'\i}a and Shawe-Taylor, John and Orlic, Davor and Kay, Jackie and Siegel, Noah and Evans, Katherine and Moorosi, Nyalleng and Eliassi-Rad, Tina and Tanczer, Leonie Maria and others},
  year={2024},
  publisher={UNESCO, IRCAI}
}

@article{hamilton2017inductive,
  title={Inductive representation learning on large graphs},
  author={Hamilton, Will and Ying, Zhitao and Leskovec, Jure},
  journal={Advances in neural information processing systems},
  volume={30},
  year={2017}
}

@article{velivckovic2017graph,
  title={Graph attention networks},
  author={Veli{\v{c}}kovi{\'c}, Petar and Cucurull, Guillem and Casanova, Arantxa and Romero, Adriana and Lio, Pietro and Bengio, Yoshua},
  journal={arXiv preprint arXiv:1710.10903},
  year={2017}
}

@article{mnih2013playing,
  title={Playing atari with deep reinforcement learning},
  author={Mnih, Volodymyr and Kavukcuoglu, Koray and Silver, David and Graves, Alex and Antonoglou, Ioannis and Wierstra, Daan and Riedmiller, Martin},
  journal={arXiv preprint arXiv:1312.5602},
  year={2013}
}

@article{zhou2023large,
  title={Large Language Model Soft Ideologization via AI-Self-Consciousness},
  author={Zhou, Xiaotian and Wang, Qian and Wang, Xiaofeng and Tang, Haixu and Liu, Xiaozhong},
  journal={arXiv preprint arXiv:2309.16167},
  year={2023}
}

@article{caliskan2017semantics,
  title={Semantics derived automatically from language corpora contain human-like biases},
  author={Caliskan, Aylin and Bryson, Joanna J and Narayanan, Arvind},
  journal={Science},
  volume={356},
  number={6334},
  pages={183--186},
  year={2017},
  publisher={American Association for the Advancement of Science}
}

@inproceedings{nangia2020crows,
    title = "{C}row{S}-Pairs: A Challenge Dataset for Measuring Social Biases in Masked Language Models",
    author = "Nangia, Nikita  and
      Vania, Clara  and
      Bhalerao, Rasika  and
      Bowman, Samuel R.",
    editor = "Webber, Bonnie  and
      Cohn, Trevor  and
      He, Yulan  and
      Liu, Yang",
    booktitle = "Proceedings of the 2020 Conference on Empirical Methods in Natural Language Processing (EMNLP)",
    month = nov,
    year = "2020",
    address = "Online",
    publisher = "Association for Computational Linguistics",
    url = "https://aclanthology.org/2020.emnlp-main.154",
    doi = "10.18653/v1/2020.emnlp-main.154",
    pages = "1953--1967",
}

@inproceedings{may2019measuring,
  title={On measuring social biases in sentence encoders},
  author={May, Chandler and Wang, Alex and Bordia, Shikha and Bowman, Samuel R and Rudinger, Rachel},
  booktitle={Proceedings of the 2019 Conference of the North American Chapter of the Association for Computational Linguistics: Human Language Technologies, Volume 1 (Long and Short Papers)},
  pages={622--628},
  year={2019}
}

@inproceedings{webster2020measuring,
  title={Measuring and reducing gendered correlations in pre-trained models},
  author={Webster, Kellie and Recasens, Marta and Axelrod, Vera and Ringgaard, Michael},
  booktitle={arXiv preprint arXiv:2010.06032},
  year={2020}
}

@inproceedings{salazar2019masked,
  title={Masked language model scoring},
  author={Salazar, Julian and Liang, Davis and Nguyen, Toan Q and Kirchhoff, Katrin},
  booktitle={arXiv preprint arXiv:1910.14659},
  year={2019}
}

@article{rajpurkar2016squad,
  title={SQuAD: 100,000+ questions for machine comprehension of text},
  author={Rajpurkar, Pranav and Zhang, Jian and Lopyrev, Konstantin and Liang, Percy},
  journal={arXiv preprint arXiv:1606.05250},
  year={2016}
}

@inproceedings{nozza2021honest,
  title={HONEST: Measuring hurtful sentence completion in language models},
  author={Nozza, Debora and Bianchi, Federico and Hovy, Dirk},
  booktitle={Proceedings of the 2021 Conference of the North American Chapter of the Association for Computational Linguistics: Human Language Technologies},
  year={2021}
}

@inproceedings{nadeem2020stereoset,
    title = "{S}tereo{S}et: Measuring stereotypical bias in pretrained language models",
    author = "Nadeem, Moin  and
      Bethke, Anna  and
      Reddy, Siva",
    editor = "Zong, Chengqing  and
      Xia, Fei  and
      Li, Wenjie  and
      Navigli, Roberto",
    booktitle = "Proceedings of the 59th Annual Meeting of the Association for Computational Linguistics and the 11th International Joint Conference on Natural Language Processing (Volume 1: Long Papers)",
    month = aug,
    year = "2021",
    address = "Online",
    publisher = "Association for Computational Linguistics",
    url = "https://aclanthology.org/2021.acl-long.416",
    doi = "10.18653/v1/2021.acl-long.416",
    pages = "5356--5371",
}

@inproceedings{gehman2020realtoxicityprompts,
  title={RealToxicityPrompts: Evaluating Neural Toxic Degeneration in Language Models},
  author={Gehman, Samuel and Gururangan, Suchin and Sap, Maarten and Choi, Yejin and Smith, Noah A.},
  booktitle={Findings of the Association for Computational Linguistics: EMNLP 2020},
  pages={3356--3369},
  year={2020},
  publisher={Association for Computational Linguistics},
  doi={10.18653/v1/2020.findings-emnlp.301},
  url={https://aclanthology.org/2020.findings-emnlp.301}
}

@inproceedings{sap2020social,
  title={Social Bias Frames: Reasoning about Social and Power Implications of Language},
  author={Sap, Maarten and Gabriel, Saadia and Qin, Lianhui and Jurafsky, Dan and Smith, Noah A and Choi, Yejin},
  booktitle={Proceedings of the 58th Annual Meeting of the Association for Computational Linguistics},
  pages={5477--5490},
  year={2020},
  organization={Association for Computational Linguistics},
  doi={10.18653/v1/2020.acl-main.486},
  url={https://aclanthology.org/2020.acl-main.486}
}

@inproceedings{hessel2018rainbow,
  title={Rainbow: Combining improvements in deep reinforcement learning},
  author={Hessel, Matteo and Modayil, Joseph and Van Hasselt, Hado and Schaul, Tom and Ostrovski, Georg and Dabney, Will and Horgan, Dan and Piot, Bilal and Azar, Mohammad and Silver, David},
  booktitle={Proceedings of the AAAI conference on artificial intelligence},
  volume={32},
  number={1},
  year={2018}
}

@inproceedings{dann2022guarantees,
  title={Guarantees for epsilon-greedy reinforcement learning with function approximation},
  author={Dann, Chris and Mansour, Yishay and Mohri, Mehryar and Sekhari, Ayush and Sridharan, Karthik},
  booktitle={International conference on machine learning},
  pages={4666--4689},
  year={2022},
  organization={PMLR}
}

@book{sutton2018reinforcement,
  title={Reinforcement learning: An introduction},
  author={Sutton, Richard S and Barto, Andrew G},
  year={2018},
  publisher={MIT press}
}

@article{oketunji2023large,
  title={Large Language Model (LLM) Bias Index--LLMBI},
  author={Oketunji, Abiodun Finbarrs and Anas, Muhammad and Saina, Deepthi},
  journal={arXiv preprint arXiv:2312.14769},
  year={2023}
}

@techreport{hagberg2008exploring,
  title={Exploring network structure, dynamics, and function using NetworkX},
  author={Hagberg, Aric and Swart, Pieter J and Schult, Daniel A},
  year={2008},
  institution={Los Alamos National Laboratory (LANL), Los Alamos, NM (United States)}
}

@article{paszke2019pytorch,
  title={Pytorch: An imperative style, high-performance deep learning library},
  author={Paszke, Adam and Gross, Sam and Massa, Francisco and Lerer, Adam and Bradbury, James and Chanan, Gregory and Killeen, Trevor and Lin, Zeming and Gimelshein, Natalia and Antiga, Luca and others},
  journal={Advances in neural information processing systems},
  volume={32},
  year={2019}
}

@article{tan2024large,
  title={Large Language Models for Data Annotation: A Survey},
  author={Tan, Zhen and Beigi, Alimohammad and Wang, Song and Guo, Ruocheng and Bhattacharjee, Amrita and Jiang, Bohan and Karami, Mansooreh and Li, Jundong and Cheng, Lu and Liu, Huan},
  journal={arXiv preprint arXiv:2402.13446},
  year={2024}
}

@article{zhang2023llmaaa,
  title={Llmaaa: Making large language models as active annotators},
  author={Zhang, Ruoyu and Li, Yanzeng and Ma, Yongliang and Zhou, Ming and Zou, Lei},
  journal={arXiv preprint arXiv:2310.19596},
  year={2023}
}

\end{document}